\documentclass[runningheads]{llncs}
\usepackage{graphicx}
\usepackage{times}
\usepackage{epsfig}
\usepackage{graphicx}
\usepackage{amsmath}
\usepackage{amssymb}
\usepackage{algorithm}
\usepackage{subcaption}
\usepackage{hhline}
\usepackage{url}
\usepackage[noend]{algpseudocode}
\usepackage[dvipsnames]{xcolor}

\newcommand*\Let[2]{\State #1 $\gets$ #2}
\DeclareMathOperator*{\argmax}{arg\,max}
\begin{document}
\title{Color inference from semantic labeling for person search in videos}
%
%
\author{Jules Simon\inst{1}\and
Guillaume-Alexandre Bilodeau\inst{1} \and
David Steele\inst{2} \and Harshad Mahadik \inst{2}}
\authorrunning{J. Simon et al.}
%
\institute{LITIV lab., Polytechnique Montréal, Montréal QC, H3T 1J4, Canada\and
Arcturus Networks, Etobicoke, ON, M9C 1A3, Canada \\
\email{\{jules.simon, gabilodeau\}@polymtl.ca, \\ \{dsteele, harshad\}@arcturusnetworks.ca}
}
\maketitle              

\setcounter{footnote}{0}
\begin{abstract}
    We propose an explainable model for classifying the color of pixels in images.
    We propose a method based on binary search trees and a large peer-labeled color name dataset, allowing us to synthesize the average human perception of colors.
    We test our method on the application of Person Search.
    In this context, persons are described from their semantic parts, such as \textit{hat, shirt, \ldots} and person search consists in looking for people based on these descriptions.
    We label segments of pedestrians with their associated colors and evaluate our solution on datasets such as PCN and Colorful-Fashion. We show a precision as high as 83\% as well as the model ability to generalize to multiple domains with no retraining.

\end{abstract}
\section{Introduction}

Security matters are prevalent today, and so does the use of video surveillance.
Therefore, efforts are being made to develop automatic methods for labeling videos and searching inside them for people or events. 
Given bounding boxes of people in images, we aim to generate labels that can be used for person search in videos using colors and semantic parts. Our goal is to answer queries such as: \textit{Find in these videos a person with a red shirt and blue pants.} The answer to such query can then be used to obtain or filter candidates for person re-identification. 

Within our context of person search, colors are an efficient way of describing people.
As the images produced by cameras used in surveillance often suffer from defects such as low resolution, dirty lenses, haze and video compression artifacts, we can use colors as a discriminative feature to easily filter out true negatives from a query.

\begin{figure}[H]
    \centering
    \includegraphics[width=0.7\textwidth]{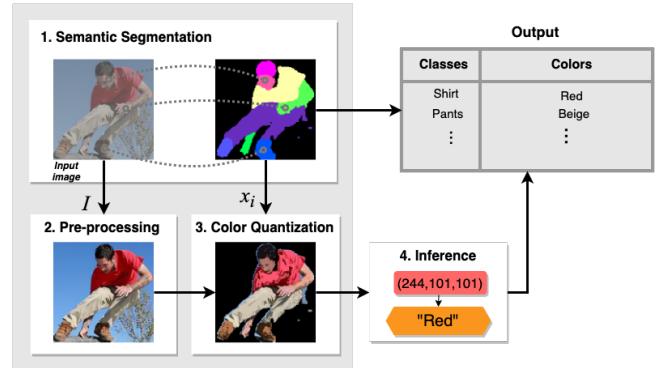} 
    \caption{Overview of the main steps of our method: Segmentation, color pre-processing, quantization and classification.}
    \label{method}
\end{figure}

Thus, we are interested in classifying colors accurately in images. This task is not often the main focus of computer vision works and for instance, most datasets, such as Colorful-Fashion Parsing \cite{liuFashionParsingWeak2014a}, DukeMTMC-attribute \cite{zheng2017unlabeled} and Market-1501-attribute \cite{1501} only provide 8 to 13 discrete color attributes in their labels.
A richer color model would enable the generation of more organic and convincing textual descriptions of images, and fit these descriptions closer to human perception, making natural language queries more streamlined for person search.

Since color labels are meant to be used by humans, we ask ourselves the following question: 
\textit{How can we generate meaningful, non-ambiguous semantic color labels for describing persons ?} 
Trying to answer this question calls for the following requirements: 1)  we need a model that reflects the human visual system,
2) we need an appropriate way of sampling colors in images in order to associate them to semantic classes, and 3) we have to make sure that good classification results translate well into the real-world application of person search.

We aim to satisfy these requirements with the following contribution: we introduce a new method for generating semantic labels of persons in videos using colors. It uses semantic segmentation \cite{MacroMicroAdversarial2018luo} in order to associate semantic meaning to pixels in the images. 
Images are pre-processed and the colors of each semantic part are quantized in order to extract their dominant color. Finally, color classification is performed on these dominant colors. 
Figure \ref{method} gives an overview of our method. 


\section{Background and related work}

\subsection{Color perception}
To work on color label inference, we first try to understand how the brain perceives colors and how to emulate this perception with a computer.
According to \cite{ComputerVisionszeliski} our visual system's perception of colors is intrinsically non-uniform and thus calls for more flexible models. 
In addition to this property of our vision, as highlighted by \cite{ColorPers}, the perceived color of an object varies with the properties of its materials and its environment. To describe formally this phenomenon, a given pixel in an image can be described as the product of its illuminance (the quantity of light hitting a surface) and its reflectance (the quantity of light reflected of a surface).
By separating the illuminance and reflectance information, our visual system perceives colors consistently regardless of the illumination through a mechanism called "color constancy".

In order to process an image such that its colors are represented as our visual system would see them, and bridge the gap between the colors we see and the ones we experience, the popular algorithm Retinex \cite{Lightnessretinex1971land} can be used. 
A lightness-color constancy algorithm such as Retinex must achieve three tasks \cite{Propertiesperformance1997jobson}: 1) compress the dynamic range of the scene, 2) make colors independent of the illumination of the scene and 3) keep the color and lightness rendition.
Improvements over the original Retinex algorithm allow for more general use cases, 
for example, the Automated Multiscale Retinex with Color Restoration \cite{MultiscaleRetinex2014petro} is an image independent Retinex algorithm that solves the issue of graying out (images becoming desaturated after Retinex).

Color constancy is still an ongoing problem to solve, and there are some more recent works on the subject \cite{CNNBased2018baslamisli} \cite{QuasiUnsupervisedColor2019bianco} using advances in machine learning. However only \cite{CNNBased2018baslamisli} is trained to handle lightness constancy, and \cite{QuasiUnsupervisedColor2019bianco} achieves only color constancy, i.e. the second requirement of lightness-color constancy.

\subsection{Color inference}
There are already existing methods for color inference, working mostly on a one-to-one association of color to label.
There are a number of color names lookup tables built by committees available for use such as the ISCC–NBS System of Color Designation \cite{ISCCNBSmethod1955kelly} or the X11 color names \cite{X11Color}.
These tables however feature a low number of unique names, for instance the X11 color table is 783 lines long, and moreover while some of these tables are bigger (the ISCC-NBS table features 42,000 points) they do not contain descriptive names.
In addition, color lookup tables only feature one RGB triplet per color name and therefore require interpolation to cover the whole spectrum. 
Considering that the human perception of color is not linear, this interpolation is non-trivial to define.
Finally, color tables do not capture the notion of clusters and ambiguity in perception: color names do not have variances, cluster shapes, cluster sizes and cannot overlap.

This observation comes from Mojsilovic \cite{computationalmodel2005mojsilovic} when proposing a computational model for color naming.
Starting from the ISCC-NBS \cite{ISCCNBSmethod1955kelly} dictionary, this model uses three color naming experiments in order to define a more accurate color vocabulary.
The model proposed in \cite{computationalmodel2005mojsilovic} works with 267 named color points spread around the color spectrum and allows the generation of color names for any input color. However, these names follow a strict pattern and are only abstract names with modifiers such as \textit{light, vivid, etc}. Thus, while this work is related to our problem, it does not handle semantics and ambiguity in color names (i.e. when several names could be applied to an input color).

PCN-CNN \cite{PedestrianColor2017cheng} is a more recent and comprehensive method for pedestrian color naming, based on the VGG convolutional neural network, and it is able to achieve state of the art performance for pixel-level color naming.
However, like many CNN, it has to be trained on domain-specific datasets and therefore is not a general solution for color naming.

\section{Problem statement}
Person search requires a textual description of a person. Therefore, for a given image of a person $I$, we want to generate $k$ color name labels $l_i$ for each of the $k$ semantic classes present in the image.
These classes can be for instance \textit{shirt, shoes, scarf, etc}, and the color labels should be meaningful and reflect the many different possibilities for naming colors based on perception.

\section{Method}
We focus our work on a data-driven model for color naming.
Our algorithm has four main steps as highlighted in figure \ref{method}, and we detail them in the following sections.
In a first step, we perform semantic segmentation on a pedestrian image to extract binary masks corresponding to different body parts.
In a second step, we process the image to enhance its colors, then we quantize them within each semantic mask.
Finally, we perform color classification using the semantic binary masks computed in the first step and the image computed in the third step.
As an output, we generate for a given image of a pedestrian a table associating semantic classes and colors.

\subsection{Semantic segmentation}
Let $I$ be an input image consisting of a crop of a person.
As a first step, we compute the semantic segmentation of $I$ for $k$ classes: $M = \phi(I)$, where $M$ is a tensor of $k$ two-dimensional binary masks.
They correspond to the semantic parts of the person such as \textit{torso, legs, feet, etc.} from which we can sample colors.

The semantic segmentation and labelling is achieved with a GAN, the MMAN architecture \cite{MacroMicroAdversarial2018luo}, that we selected for its good ability to generalize.
We train it on the PPSS dataset \cite{Pedestrianparsing2013luo} as it features images of people in low contrast situations that correspond to a worst case scenario for city video surveillance footage.

As we show in our experiments in section \ref{sec:smoothing}, the quality of the semantic segmentation part is not critical to our method,  and our goal at this step is to obtain a rough localisation of body parts.

\subsection{Image pre-processing} \label{sec:preprocessing}
In this second step, we process the input images to prepare them for color classification. 
This is required as the images captured from city cameras often suffer from haze (pollution) and low contrast, as well as low saturation.
As the scene illumination is simple for street scenes in daylight (there is only one illuminant), we do not find the need for complex methods such as neural networks.

Therefore, we kept two approaches for pre-processing:
\begin{itemize}
    \item A domain specific approach, in which we search for the optimal contrast, brightness and saturation enhancements to apply to the input images using the validation set.
    \item A general approach, using the Multi-Scale Retinex with Chromacity Preservations (MSRCP) algorithm. We use the automatic method MSRCP \cite{MultiscaleRetinex2014petro} as it performs better than MSRCR \cite{automatedmulti2012parthasarathy} under even and white illumination \cite{MultiscaleRetinex2014petro}, which is often the case in outdoor scenes.
\end{itemize}

\subsection{Color quantization}
We erode each of the binary semantic masks in order to avoid border effects as the segmentation is less precise around the borders, and can introduce noise from the background or other body parts.
Then, for each mask, we quantize the colors of the underlying pixels of the pre-processed image using K-means clustering in RGB space with a small $K$, fixed to 5 in this paper. 
We then keep the biggest cluster and use its centre as the RGB color triplet to classify.

\subsection{Color classification}
To classify the colors, we use the results of the XKCD Color Surveyn that was opened in 2010 by the cartoonist Randall Munroe \cite{ColorSurvey2010munroe}.
In this survey, volunteer Internet users were shown patches of plain colors on their web browser, and for each color they were tasked with filling a free-form text box with the name they would choose for the color. 
In total 3,083,876 unique RGB tuples were assigned 183,401 unique textual color names, with a total of 3.4 million entries, that are available online.

The data extracted from this survey addresses some of the shortcomings of the aforementioned color lookup tables. For instance, it contains both abstract and descriptive labels : the abstract names allow us to capture the common  perception of the survey participants while the descriptive ones allow us to capture intrinsic semantic knowledge of the objects used for naming. For instance, labels \textit{apple, peach, sky, etc.} each have distinct color shades associated to them.
Thus, this dataset should be able to capture the variability of naming colors.
Moreover, due to its crowd-sourced nature, this dataset uses a varied vocabulary and allows to generate more realistic labels.
For the same reason, this survey was filled using different screen technologies and thus is closer to the variance in color perception that would happen in real-world scenarios. 
However, there is a noteworthy problem introduced by the free-form text boxes used for the survey, which is the introduction of noise in the labels as participants can input any answers, including irrelevant ones. Thus we need to remove these outliers. 

Using the results of the survey, we classify the RGB triplets using a binary decision tree. 
Our choice of a decision tree is motivated by the fact that it does not use distances but instead binary comparisons, making it efficient regardless of the color spaces.
Moreover, decision trees can be used for multi-label classification and thus allow us to deal better with ambiguity by outputing several possible color names for a given pixel value.
Finally, a tree model has the advantage of being explainable and easy to visualize, even more so with color values as decision variables, and is fast at inference time.

\subsubsection{Decision Tree training}

During the first filtering step in line \ref{algo:line} of algorithm \ref{algo:tree}, we sort the color names dataset by label occurrences and only keep the most common color names, such that only a given portion $\tau$ of the dataset is represented. This ratio $\tau$ allows us to filter out names that are too unique to be representative.
We find that in our situation, because the labels were generated using a free-form survey, $\tau = 0.65$ is a good choice as this corresponds to labels having more than 2000 occurrences and therefore representative of a consensus. This gives a dataset of 140 unique labels and 2,263,631 samples. This data is available to download \footnote{The data is available at \url{https://github.com/Smoltbob/XKCDColors-Dataset}.}.

\begin{algorithm}
  \caption{Building the decision tree}
  \label{algo:tree}
  \begin{algorithmic}[1]
    \Require{$D$ color names dataset}
    \Require{$L$ the set of colors labels on which to train}
    \Statex
    \Statex Sort $D$ by occurences
    \Let{$D'$}{MostFrequentLabels($D$, $\tau$)} \label{algo:line}
    \Let{$D''$}{RemoveOutliers($D'$)}
    \Let{$D'''$}{Resample($D''$)}
    \Let{$D^*$}{\begin{math}\{d \| d_{label} \in L \: \forall\: d \in D'''\}\end{math}}
    \Let{$T$}{DecisionTree($D^*$)}
  \end{algorithmic}
\end{algorithm}

Following this step (at line 2), we perform class-wise outliers removal using K-Nearest Neighbors \cite{EfficientAlgorithmsramaswamy}.
This step allows us to remove any data point that has less than $K$ neighbors and is used to make the convex hull of each color cluster smoother, which is useful for the following re-sampling step as well as to reduce ambiguity between small clusters of colors.

In the next step (line 3), re-sampling is done using  Synthetic Minority Over-sampling Technique (SMOTE \cite{SMOTEsynthetic2002chawla}).
This oversampling algorithm works by generating new samples within the convex hull of the minority classes. This allows us to balance the dataset as some colors are underreesented.

In the end we obtain $D^*$ (line 4), the final dataset from which we can train the tree $T$ (at line 5) with the subset of labels $L$.
We train it on a subset of prototype color names, such as the 11 colors \textit{(black, white, red, green, yellow, blue, brown, pink, orange, purple, and gray)} as defined by Berlin and Kay \cite{BasicColor1999berlin}. This choice can also be made according to the domain of the application at hand.

\subsubsection{Pooling images in videos} \label{sec:pooling}
We can make use of the additional information offered in a video by sampling using several frames.
For a given pedestrian, sampling on several frames allows us to reduce the uncertainty and be more robust to noisy events that alter color rendition such as walking in the shade.

If several images are available for a given person, we can use them for the classification using one of three following pooling methods: 1) randomly using one image per person, 2) averaging the colors per mask for all of the images of each person or 3) sorting the colors by saturation and classifying on the most saturated images.

\section{Experiments} \label{sec:experiments}
In order to evaluate performance for person search, we use the Region Annotation Score (RAS) \cite{PedestrianColor2017cheng}. This score is a region-wise metric, computed over the dataset using region labels. It is equivalent to the Precision for regions.
We compute it using \[ RAS = \frac{TP}{TP+FP} \] where $TP, FP$ are the number True and False positives and $FN$ are the number of False negatives.
A true positive corresponds in practice to a successful retrieval of an attribute for a given query.
We use this metric as it allows us to compare our methods to the benchmark of \cite{Pedestrianparsing2013luo}. However, we also include a study of the recall in section \ref{sec:smoothing}.
We performed our experiments using MSRCP pre-processing on the images. For training, we used SMOTE resampling and outlier removal. Finally, we used average pooling when computing the final classification result.

\subsection{Datasets}

The first dataset used is the Pedestrian Color Naming (PCN) Dataset \cite{PedestrianColor2017cheng}. It is a color-balanced split of the Market-1501 dataset \cite{1501}. Market-1501 is made for person re-identification and features 1501 identities captured in 32,668 bounding boxes. The images in this dataset are similar to a real world situation, with low quality street scenes.

PCN is augmented by pixel-level color annotation maps, rather than image-level as in Market-1501-Attribute. We used it to compare our method with the other methods that were tested in \cite{PedestrianColor2017cheng} on this dataset.
We followed the same procedure as described in the original paper: we use the dataset ground truth, and we measure the RAS for color prediction.

We also used the Colorful-Fashion dataset \cite{liuFashionParsingWeak2014a}, that  focuses on women fashion and provides clothes description for each of its 2682 images, semantic segmentation masks and a color for each class.
The dataset is generated automatically with SLIC superpixel segmentation \cite{Slicsuperpixels2010achanta}.
With this dataset, we used the provided masks instead of computing new ones, as they have more granular classes.
As this dataset is vivid and well lit, we did not apply any pre-processing on it and used the images as is.

\subsection{Results and discussion}
Results are reported in table \ref{tab:comparison}. 
For our method, we used two training settings: in a first setting, we used the model trained on XKCD as described in the method section, and in a second setting, we retrained the model on the pixel distributions of the datasets. 
This is done in order to compare our method with methods such as PCN-CNN \cite{PedestrianColor2017cheng} or SVM \cite{liuFashionParsingWeak2014a}, that are trained on specific datasets as well.

We show that with retraining, our model achieves its best performance on both dataset and gives results close to the state of the art, and that without retraining it remains close to the state of the art, losing about 6 to 9 points in precision.

\begin{table}[h]
\caption{Classifier RAS study. \textcolor{red}{\textbf{Boldface}} indicates best results, \textcolor{blue}{\textit{italic}} second best.} 
\centering
\begin{tabular}{c|c|c|}
\hline
\multicolumn{1}{|c|}{Method} & \multicolumn{1}{c|}{PCN} & \multicolumn{1}{c|}{Colorful-Fashion} \\ \hline\hline
\multicolumn{1}{|c|}{PLSA \cite{LearningColor2009weijer}} &  68.4 & 71.4 \\
\multicolumn{1}{|c|}{PFS \cite{Parametricfuzzy2008benavente}} & 68.5 & 60.5 \\
\multicolumn{1}{|c|}{SVM \cite{liuFashionParsingWeak2014a}}  & 62.2 & 45.4 \\
\multicolumn{1}{|c|}{DCLP \cite{IlluminationRobust2015liu}} & 62.0 & 54.8 \\ 
\multicolumn{1}{|c|}{PCN-CNN \cite{PedestrianColor2017cheng}} & \textcolor{red}{\textbf{80.8}} & \textcolor{blue}{\textit{81.9}} \\
\multicolumn{1}{|c|}{Ours}  & 75.0 & 76.1 \\
\multicolumn{1}{|c|}{Ours, retrained}  & \textcolor{blue}{\textit{80.4}} & \textcolor{red}{\textbf{83.0}} \\
\hline
\end{tabular}
\label{tab:comparison}
\end{table}

\section{Study of the model} 
In this part, we look into the limits of our model as well as into ways of optimizing it.

We used the Market-1501-Attributes dataset \cite{1501_att}, that is built from the Market-1501 dataset \cite{1501}.
Market-1501 Attributes \cite{1501_att} is an augmentation of this dataset with 28 hand annotated attributes, such as gender, age, sleeve length, flags for items carried, as well as upper clothes colors and lower clothes colors. Here we used the upper clothes or lower clothes labels, that we matched with the classes of the same name from the PPSS dataset.
We selected this dataset to test the whole person search pipeline as described in our methodology going from the segmentation to the classification of colors.

Results in this section correspond to the average for 10,000 picture queries over 500 identities from the Market-1501 Attributes dataset \cite{1501_att}. 

\subsection{Evaluation of pre-processing approaches} \label{sec:processing}
To facilitate the distinction between colors and improve the person search results, we can pre-process the input images, as described in \ref{sec:preprocessing}.  In this experiment we compared the two approaches (using learned hyper parameters or using MSRCP) to a baseline.  

We also compared the three pooling methods of images mentioned in section \ref{sec:pooling} : random sampling, average pooling and saturation sort (Sat Sort), applied on the dataset. We measured the RAS for the upper clothes and the lower clothes. Results are reported in table \ref{tab:market}.

\begin{table}[h]
\caption{RAS for Market-1501 under different pre-processing configurations. \textcolor{red}{\textbf{Boldface}} indicates best results, \textcolor{blue}{\textit{italic}} second best.} 
\centering
\begin{tabular}{c|c|c|c|}
\cline{2-4}
\multicolumn{1}{c|}{} & \multicolumn{1}{c|}{None} & \multicolumn{1}{c|}{Learned} & \multicolumn{1}{c|}{MSRCP} \\
\hhline{-===} 
\multicolumn{1}{|c|}{Random} & 69.6  & 68.6 & 65.2 \\ \hline
\multicolumn{1}{|c|}{Average} & 69.1 & \textcolor{red}{\textbf{74.4}} & 72.3 \\ \hline
\multicolumn{1}{|c|}{Sat Sort} & 69.7 & \textcolor{blue}{\textit{73.1}} & 71.2 \\ \hline
\end{tabular}
\label{tab:market}
\end{table}

For the average pooling of images, we notice that using the parameters learned from the data improves the RAS by 5 points compared to the baseline. However this pre-processing method is costly and domain dependant and we see that using a method independent of the domain such as MSRCP can also improve results, by a significant margin (3 points).

\begin{figure}[h]
    \centering
    \begin{subfigure}[t]{.48\columnwidth}
        \centering
        \includegraphics[width=0.7\textwidth]{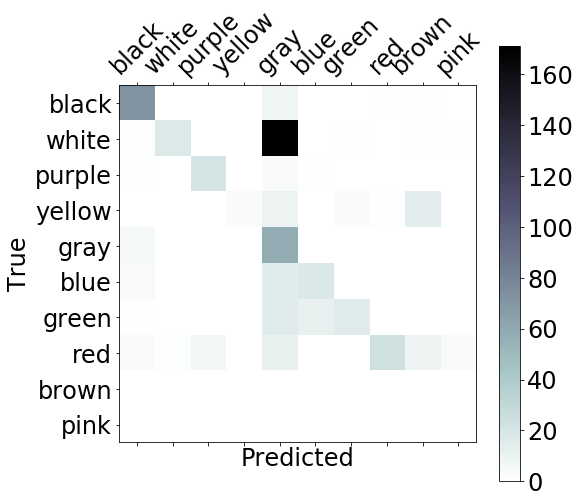} 
        \caption{No pre-processing.}
    \end{subfigure}
    \begin{subfigure}[t]{.48\columnwidth}
        \centering
        \includegraphics[width=0.7\columnwidth]{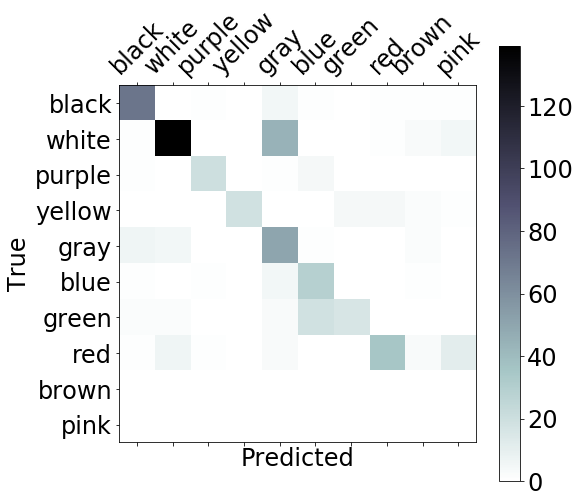}
        \caption{MSRCP pre-processing.}
    \end{subfigure}
    \caption{Confusion matrices for predictions in Market-1501, with and without pre-processing the images.}
    \label{confusion}
\end{figure}

As shown in figure \ref{confusion}, with no pre-processing, most of the failure cases concern shades of white and the pre-processing step helps alleviate this issue.

Compared to the saturation sort, the simple method of average pooling gives a better overall result and has a lower processing time
We hypothesize that while saturation sort increases the performance on vivid colors, it does so at the cost of reduced performance on neutral colors and makes them more sensitive to color noise and compression artifacts.
Random pooling, which uses a single image, by far gives the worst results. 
This shows that in situations where several images of the same object are available, such as in videos, we can noticeably improve the performance by adding a pooling step.

\subsection{Training strategy} \label{sec:training}
Using average pooling and MSCRP pre-processing, we compare our different training data processing steps using the XKCD dataset as synthetic data as well as the Market-1501 dataset.
These steps can consist of removing outlier points (Clean) or applying SMOTE resampling to balance the dataset. Results are given in table \ref{tab:raw}.

\begin{table}[H]
\caption{Model performance study. \textcolor{red}{\textbf{Boldface}} indicates best results, \textcolor{blue}{\textit{italic}} second best. Synthetic is the XKCD test set.} 
\centering
\begin{tabular}{cc|c|c|}
\hline
\multicolumn{1}{|c}{SMOTE} & Clean & \multicolumn{1}{c|}{Synthetic} & \multicolumn{1}{c|}{Market} \\ \hline\hline
\multicolumn{1}{|c}{} &  & 87.6 & 72.3 \\
\multicolumn{1}{|c}{} & \checkmark & \textcolor{blue}{\textit{97.5}} & \textcolor{red}{\textbf{74.9}} \\
\multicolumn{1}{|c}{\checkmark} &  & 89.2 & 72.3 \\
\multicolumn{1}{|c}{\checkmark} & \checkmark & \textcolor{red}{\textbf{97.7}} & \textcolor{blue}{\textit{73.7}} \\ \hline
\end{tabular}
\label{tab:raw}
\end{table}

In Market-1501, results do not vary much regardless of the training parameters, while there is up to a 10-point difference between combinations when classifying the synthetic data. 
We explain this difference by the non-uniform distribution of colors: in Market-1501, the most common colors are \textit{green, blue, purple} and the least common are \textit{white, black, gray}. These last colors are the hardest to classify because of the color constancy problem, which does not occur with synthetic data.

\subsection{Robustness of the method to segmentation maps} \label{sec:smoothing}
In this experiment, we show that our method is robust to the quality the segmentation maps it is provided with, and that it can achieve good results even when the segments are altered. To do so, we introduce an operation that we named "Semantic Smoothing". 
For a given semantic map $M$ made of $k$ labels, we computed its smoothed version $M'$ with the equation \ref{equation}.
\begin{equation} \label{equation}
M'(x,y) = \argmax_{k}[\sum_{s=-h}^{h} \sum_{t=-h}^{h} w(s,t, \sigma) * M_k(x-s, y-t)]
\end{equation}
where $w(x,y, \sigma)$ is a Gaussian kernel of size $h$ and $M_k$ the semantic map of label $k$.

\begin{figure}[H]
    \centering
    \begin{subfigure}{.2\columnwidth}
        \centering
         \includegraphics[width=0.8\textwidth]{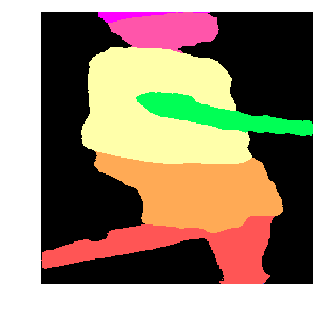}
        \caption*{$\sigma=0$}
    \end{subfigure}
    \begin{subfigure}{.2\columnwidth}
        \centering
        \includegraphics[width=0.8\textwidth]{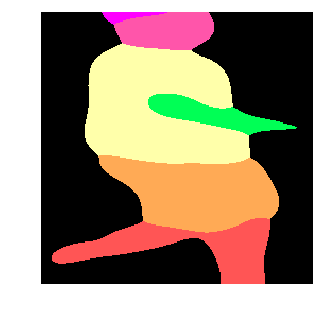}
        \caption*{$\sigma=10$}
    \end{subfigure}
    \begin{subfigure}{.2\columnwidth}
        \centering
        \includegraphics[width=0.8\textwidth]{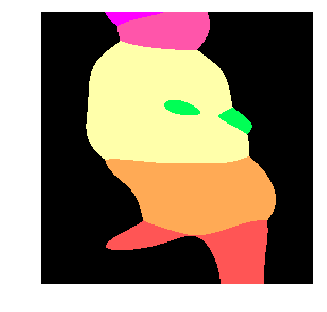}
        \caption*{$\sigma=15$}
    \end{subfigure}
    \begin{subfigure}{.2\columnwidth}
        \centering
        \includegraphics[width=0.8\textwidth]{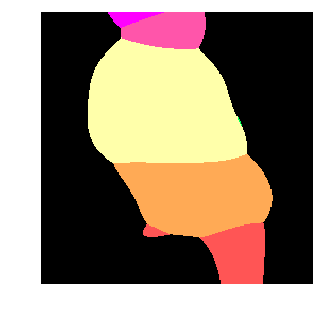}
        \caption*{$\sigma=20$}
    \end{subfigure}
    \caption{Smoothing of a semantic map for different values of $\sigma$. Small details are lost while the structure remains.}
    \label{fig:sigmas}
\end{figure}

This operation is parameterized by $\sigma$, the standard deviation of the Gaussian function.
We defined this operation as it is fast, reproducible and simulates an inaccurate semantic map as if it was outputted by a generative neural network in  a failure case. 
The main characteristics of these maps are ill-defined borders, in particular for small segments.
We show, in figure \ref{fig:sigmas}, examples of alterations of a semantic map for different values of $\sigma$.

\begin{figure}[H]
    \centering
    \includegraphics[width=0.7\textwidth]{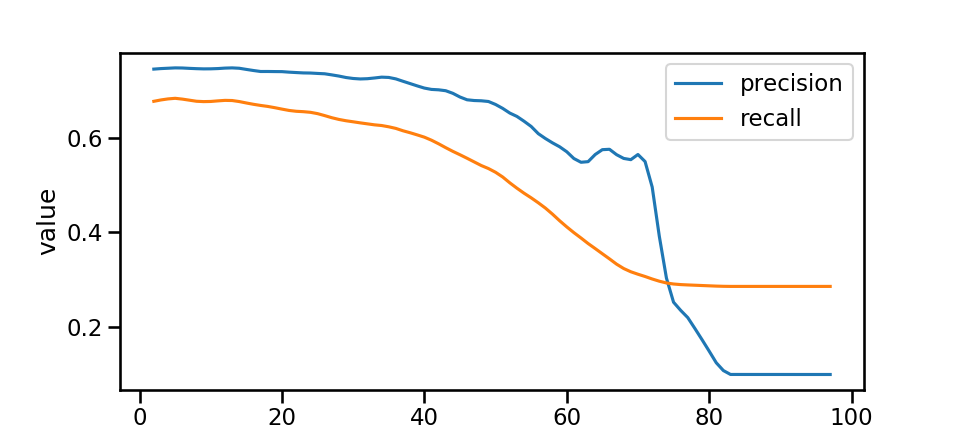}
    \caption{Evolution of the precision (RAS) and recall in function of $\sigma$ on the PCN dataset. Performance stays stable even for high $\sigma$.} 
    \label{table:sigmas}
\end{figure}

For this experiment we use the optimal parameters as described in sections \ref{sec:processing} and \ref{sec:training}. For different values of $\sigma$, we alter the semantic maps and perform person search, for the whole test set. 
We plotted the resulting measures for each sigma in figure \ref{table:sigmas}.

We notice a sharp decline where $\sigma$ reaches 60, which is the moment at which some of the bigger parts of the segments disappear.
Therefore, we conclude that full, accurate pixel-wise segmentation is not required for accurate color estimation.

\section{Conclusions}
We propose a method for generating color semantic labels from still images or videos following four main steps of 1) semantic segmentation, 2) image pre-processing, 3) color quantization, 4) color classification. 
We found that in a urban context, pre-processing the images is useful to achieve a better classification. This can be done automatically with a lightness-color constancy algorithm such as MSRCP.
We also noticed that using a crowd-sourced dataset to represent the complexity of human color vision produces results close to those of models trained using state of the art datasets such as Market-1501 or Colorful-Fashion.

We tested our method through experiments reflecting real-world situations using datasets such as PCN and Colorful-Fashion.
We achieve better results than color naming methods when retraining is not allowed. 
When allowing retraining, we show that our method achieves similar performance to state of the art deep learning solutions with respectively 80.4\% and 83.0\% of precision. This shows that we retain the same classification capabilities as state of the art trained methods while being more general domain-wise.

\paragraph{\textbf{\textup{Acknowlegments.}}} We acknowledge the support of the Natural Sciences and Engineering Research Council of Canada (NSERC), [CRDPJ 528786 - 18], and the support of Arcturus network.

{\small
\bibliographystyle{splncs04}
\bibliography{egbib}
}

\end{document}